\documentclass[letterpaper, 10pt, conference]{ieeeconf}  

\IEEEoverridecommandlockouts                              

\usepackage{graphicx} 
\usepackage{amsmath} 

\DeclareMathOperator{\argmax}{arg\,max}

\usepackage{algpseudocode}

\title{\LARGE \bf
Classifying and sorting cluttered piles of unknown objects with robots: a learning approach
}

\author{Janne V. Kujala$^{1}$, Tuomas J. Lukka$^{1}$ and Harri Holopainen$^{1}$
\thanks{$^{1}$ZenRobotics Ltd, Vilhonkatu 5 A, FI-00100 Helsinki, Finland.
        {\tt\small firstname.lastname@zenrobotics.com}}%
}

\begin{document}

\maketitle
\thispagestyle{empty}
\pagestyle{empty}

\begin{abstract}

We consider the problem of sorting a densely cluttered
pile of unknown objects using a robot.
This yet unsolved problem is relevant in the robotic waste sorting business.

By extending previous active learning approaches to grasping, we show
a system that learns the task autonomously.
Instead of predicting just whether a grasp succeeds, we predict the classes of
the objects that end up being picked and thrown onto the target conveyor.
Segmenting and identifying objects from the uncluttered
target conveyor, as opposed to the working area, is easier
due to the added structure since the thrown objects will be the only
ones present.

Instead of trying to segment or otherwise understand the cluttered working
area in any way, we simply allow the controller to learn a mapping
from an RGBD image in the neighborhood of the grasp
to a predicted result---all segmentation
etc.~in the working area is implicit in the learned function.
The grasp selection operates in two stages:
The first stage is hardcoded and outputs a distribution
of possible grasps that sometimes succeed.
The second stage uses a purely learned criterion
to choose the grasp to make from the
proposal distribution created by the first stage.

In an experiment, the system quickly learned to make good pickups and predict
correctly, in advance, which class of object it was going to pick up and
was able to sort the objects from a densely cluttered pile by color.

\end{abstract}

\section{INTRODUCTION}

\subsection{Robotic Waste Sorting}

The problem of sorting a pile of objects using a robot
is interesting in its own right,
but in our case the problem is firmly rooted in an industrial application:
waste sorting.

ZenRobotics' robots have been sorting waste on industrial waste
processing sites since 2014. Our robots have picked up
approximately 4,200 tons of metal, wood, stone and concrete
from the conveyor.  Performance of
the robots in this environment is critical for paying back the
investment. Currently the robots are able to identify, pick and
throw objects of up to 20 kg in less than 1.8 seconds, 24/7.
The current generation robot \cite{lukkazenrobotics} was taught to grasp objects using
human annotations and a
reinforcement learning algorithm.

Because of the variability of waste, the ability to recognize, grasp and
manipulate an extremely wide variety of objects is crucial.
In order to provide this ability in a cost-effective way, new
training methods, which do not rely on hardcoding or
human annotation are required.
For example, changing the shape of the gripper or adding degrees
of freedom might require all picking
logic to be rewritten or at least labor-intensive retraining unless the system
is able to learn to sort using the new gripper or degrees of freedom by itself.

In order to make our robots suitable for an industrial site, they are
built to be as simple and robust as possible: the robots
are built from COTS (common
off-the-shelf) parts, having only four degrees of freedom for position
and one for gripper opening.
The robot used for the experiment in this article is a prototype of our
current production version.

\subsection{The Robotic Waste Sorting Problem}

\label{sec:formalwastesorting}
In this section, we discuss the robotic waste sorting problem in a more
formal setting.
Objects that belong to various object classes arrive and are manipulated
by the robotic system into different chutes (end locations).
For each object class, the chute into which it should be placed is defined.
The waste sorting
problem for the robot
is then a multi-objective optimization problem with three criteria.
The combination of the criteria that is optimized depends
on the business case of the customer, but usually the true objective is
a monotonous nonlinear function of these three criteria.

The first criterion is the purity,
i.e., the percentage of the total weight of objects deposited into a chute
that belong to that chute.
Purity essentially determines whether the pile
of sorted objects is resalable (i.e., recyclable) or not.

The second criterion is the recovery rate,
i.e., the percentage of the total weight of objects of a class that were
deposited into the correct chute. This is of interest in cases
in which minimizing the amount of material that ends up in a landfill site
is important.

The third criterion is the throughput, i.e., the sum of objects' weights handled
by the system per unit time. The throughput affects how many systems are required
in parallel to sort a certain amount of waste.

The waste sorting problem involves a manipulation task similar
to the more studied problems of ``cleaning a table
by grasping'' \cite{rao2010grasping} and bin picking
\cite{domae2014fast,holz2014active,nieuwenhuisen2013mobile},
but also differs from them in several aspects:
\begin{enumerate}
\item It is not sufficient to be able to move the objects, they need
to be recognized as well in order to deposit each object into the correct chute.
\item Picking several objects at once is acceptable---even desirable,
provided that they are of the same class.
\item The objects are generally
novel and there is a large selection of different objects. Objects
can be broken irregularly.
The distribution of the objects has a long tail and completely
unexpected objects
occasionally appear.
\item The objects are placed
  on the conveyor belt by a random process and easily form
  random piles.
\item On the other hand, this problem is made slightly easier by the
fact that it is not necessary to be gentle to the objects; fragile objects will likely
have been broken by previous processes already. Scratching or colliding
with objects does not cause problems as long as the robot itself
can tolerate it (see Fig.~\ref{fig:gripper}).
\end{enumerate}

\subsection{Related work}

State-of-the-art results for grasping novel objects
and grasping objects in unstructured environments such as in dense clutter
generally use machine learning to evaluate and choose the best among
possible grasps \cite{kopicki2016one,ten2015using,katz2014perceiving}.

A recent trend is making the robot learn a grasping
task completely autonomously, using active
learning.
In order to achieve this,
the robotic system must generate the feedback data autonomously.
Grasps are automatically annotated as success or failure
based on sensors in the gripper
\cite{DBLP:journals/arobots/NguyenK14,levine2016learning,pinto16supersizing} or
visual feedback, e.g., by comparing images of the table before and after dropping
a supposedly grasped object \cite{levine2016learning}.

Once automatic feedback has been implemented, much larger amounts
of training data can be obtained than before and
the robotic system can be adapted to different
circumstances simply by letting
it learn the required behaviour in the new circumstances.

\subsection{Contributions}

Our main contribution is an autonomous robotic system
that is able to sort a densely cluttered pile of objects by class,
provided there is a way to recognize an object's class
in an uncluttered environment.

The specific novel improvements to the current state-of-the-art
models that learn to grasp or move objects are
\begin{itemize}
\item
We make no attempt to explicitly segment or understand the objects / classes
of objects in the working area.
\item
In addition to the grasp success probability, the machine learning
model is taught to predict the class distribution of the grasped objects
(enabling sorting).
\item
Feedback about object classes is obtained automatically from a more
structured environment after the robotic manipulator has grasped and
thrown the object. This makes it possible to generate a large amount of
labeled data about the cluttered environment.
\item
As the first, fixed-function stage of the system, we present an efficient
algorithm for finding \emph{closed} grasps for a two-fingered gripper from
a heightmap.
\end{itemize}

\section{PROPOSED SYSTEM FOR SORTING CLUTTERED PILES}

\begin{figure}[tb!]
  \centering
  \includegraphics[trim=250px 470px 250px 160px, clip=true, width=\columnwidth]{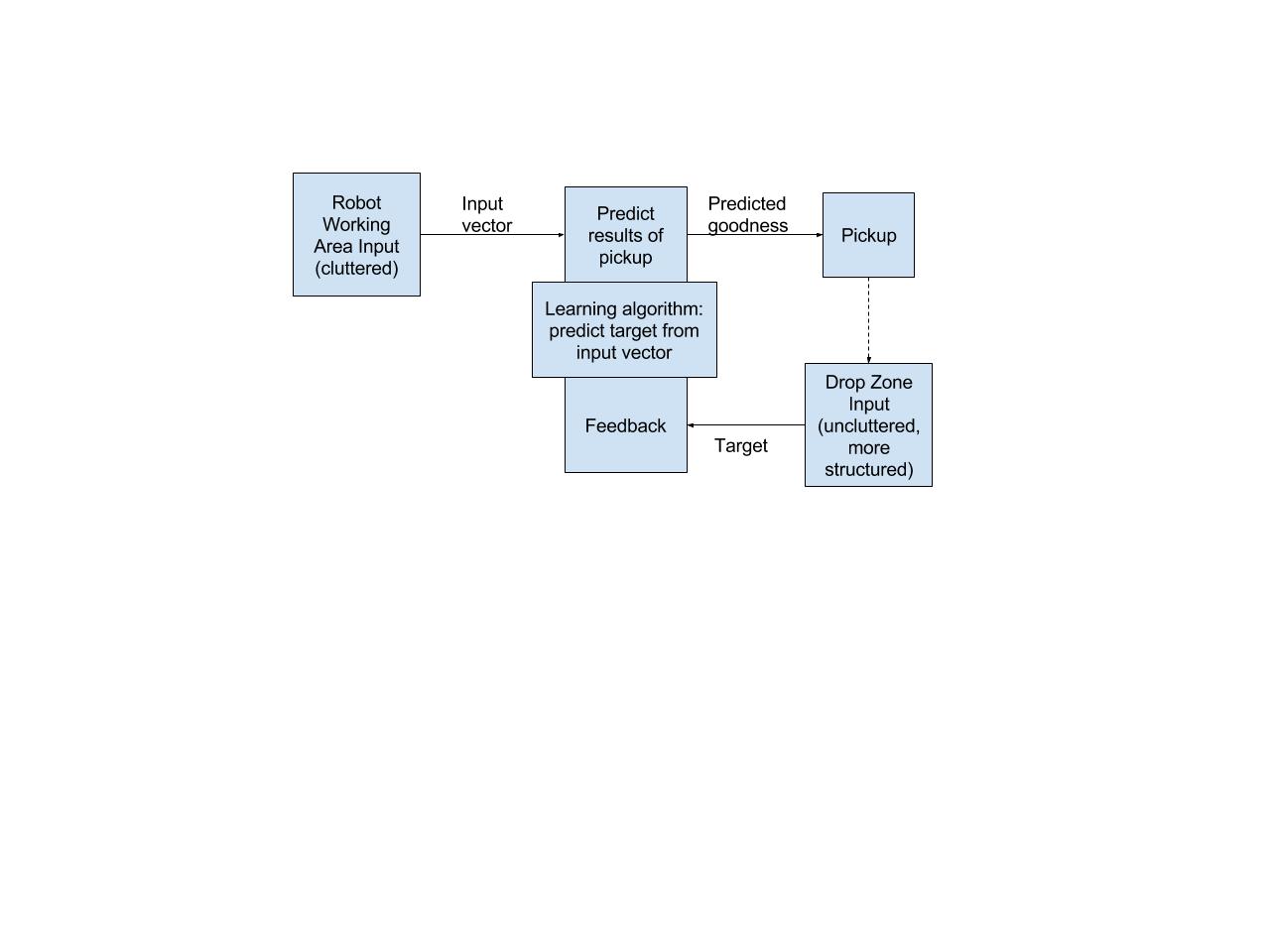}
  \caption{The proposed setup for learning to sort cluttered piles.
Feedback for learning is gathered from the actual objects thrown to the
second conveyor (``drop zone'').
At that point, classifying the objects
is much simpler and the system can see which objects actually got picked up
by its action, without interference from other objects or the gripper.
  \label{fig:overallsystem}
    }
\end{figure}

The overall architecture we propose is shown in Fig.~\ref{fig:overallsystem}.
The change from existing
systems \cite{DBLP:journals/arobots/NguyenK14,levine2016learning,pinto16supersizing}
is that instead of using a single binary grasp success-failure
feedback using, e.g., the gripper force
sensors,
we use the robot to move and drop
the grasped objects to a different area
in which we can generate richer feedback by identifying them.

\begin{figure}[tb!]
  \includegraphics[trim=140px 150px 50px 160px, clip=true, width=\columnwidth]{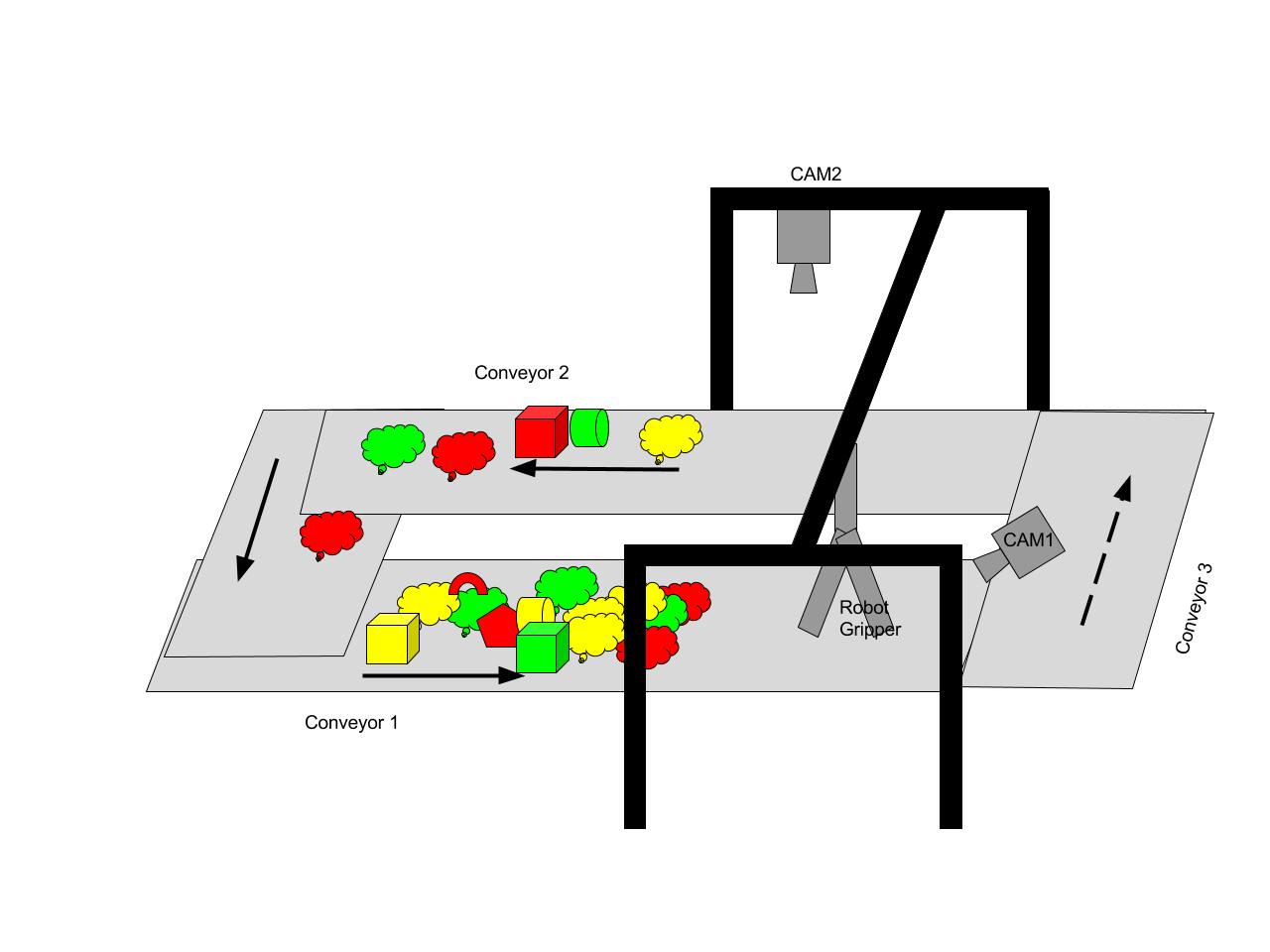}
  \caption{A diagram of the hardware used in the experiment.
    The conveyors form a closed loop to allow the test objects to be cycled
    through the system.
    During most of the operation, the conveyor shown on the right,
    which runs from the working area
    to the drop zone, is turned off to avoid unpicked objects from
    entering the drop zone. When the side conveyor has a large enough
    pile on it, the system is stopped and that conveyor is turned on to
    return the unpicked objects to the material cycle.
  \label{fig:hw}
}
\end{figure}

After dropping the object(s), the system takes a picture to generate the feedback of how much
of each class of object was visible at the drop area. After this, the drop area
is cleared for the next attempt (in our example system, the dropped objects are taken away by
a conveyor).

This architecture fulfills the requirements in \cite{DBLP:journals/arobots/NguyenK14} for
learning behaviours (i.e., 1. choice of behaviours, 2. reliable feedback on whether
a behaviour
was successful, 3. complementary behaviour: returning the world to the original state after
a behaviour, and 4. parameters for complementary behaviour from chosen behaviour).
However, the drop area being cleared is, instead of a complementary behaviour,
a \emph{nullifying} behaviour that brings the system back to
its original state without any parameters.

\section{EXPERIMENT}

\subsection{Test problem}

To simplify the recognition part of the system, we used color as a proxy for a more complete
recognition system.
The task for the system was to sort the objects into three classes: red, yellow and blue-green.
We spray-painted a number of real and simulated
waste objects with bright colors.
Both the task and the painting were chosen to make recognizing
the objects at the drop zone as simple as possible.

The weights of the objects ranged from under 100~g for light-weight plastic objects
to over 4~kg for large pieces of concrete.
Many of the objects such as the heaviest objects
 were purposefully selected so as to be difficult for the gripper being used, in order
to present a challenge for the system.

\subsection{Hardware}

The overall hardware configuration of the experiment is shown in Fig.~\ref{fig:hw}.

\subsubsection{Gantry-type robot}

We used a 4-DOF gantry-type robot with Beckhoff servos for positioning.
The degrees of freedom are three translations and a rotation around
the vertical axis.

\subsubsection{Gripper}

\begin{figure}[tb!]
  \centering
  \vspace{4pt}
  \includegraphics[trim = 400px 280px 490px 240px, clip=true, width=\columnwidth]{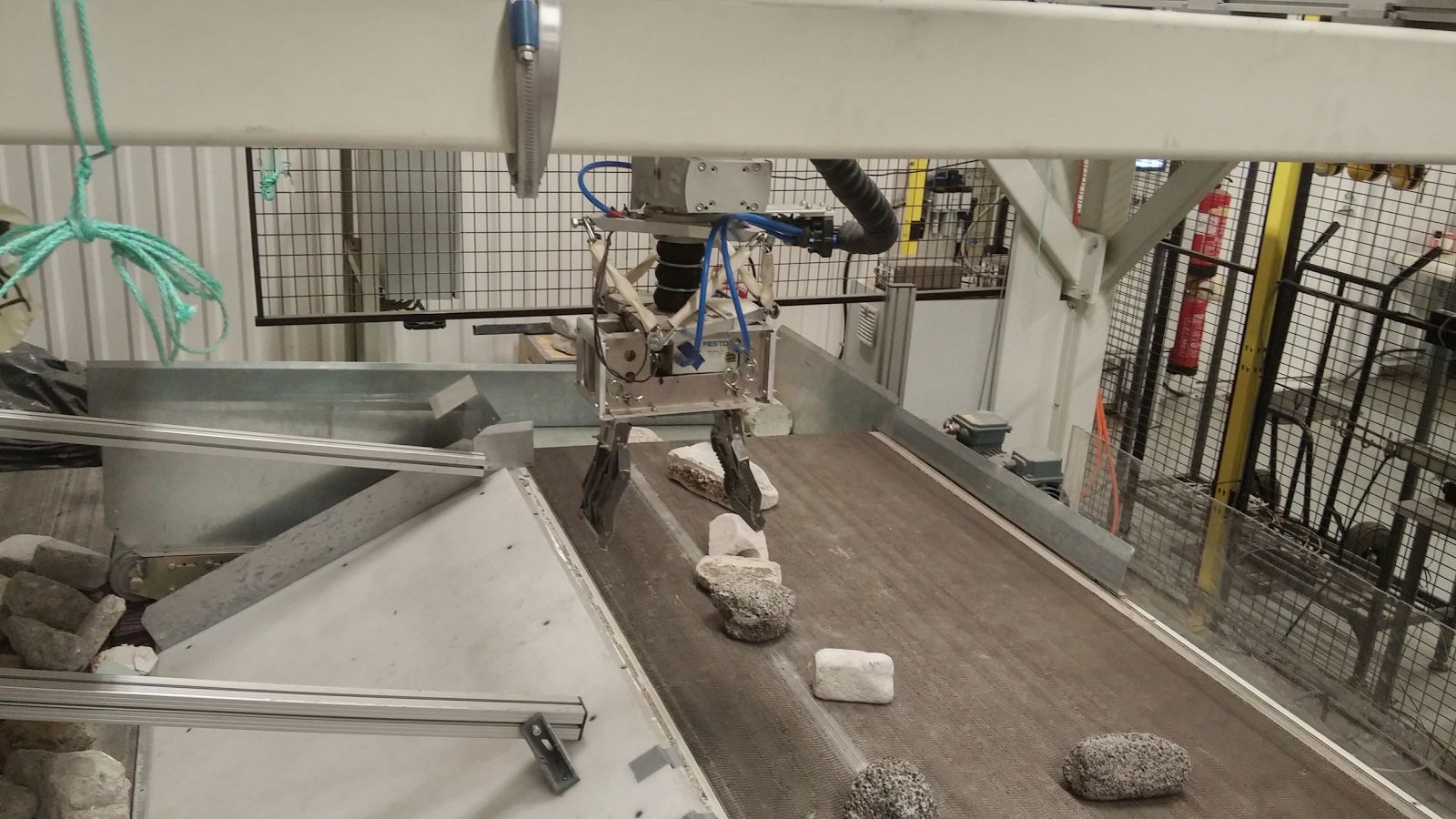}
  \caption{The gripper used in the experiments is an earlier version
    of our commercial gripper. This pneumatic gripper has a wide
    opening, is position-controllable, and contains a
    large-angle large-displacement compliance system while still being
    rigid when forces and torques do not exceed a threshold.
  \label{fig:gripper}
  }
\end{figure}

The gripper used has a wide opening and a large-angle
compliance system (Fig.~\ref{fig:gripper}).  The gripper has evolved
in previous versions of our product step by step to be morphologically
well-adapted to the task.
The gripper is pneumatically position-controllable and has a sensor giving its
current opening.

\subsubsection{Conveyor merry-go-round}

For cycling the objects through the system, we used the waste merry-go-round
depicted in Fig.~\ref{fig:hw}. This system makes it possible to run
long training sessions with a large number of objects.

Due to the merry-go-round, the system does not actually place the sorted objects
in different places by default---but it makes a prediction before throwing
an object. In a separate test (see accompanying video), we made the system successfully
throw the objects it predicted to be red
to the other side of the conveyor to ensure it is able to really sort objects.


\subsubsection{RGBD cameras}

Both the working area camera and the drop zone camera were Microsoft Kinect One time-of-flight
RGBD cameras.

\subsection{Software}

\subsubsection{Main sorting loop and data generation}

The overall algorithm of the sorting loop is described in Fig.~\ref{fig:datagen}.
This component is responsible for evaluating the situation in the working area and
carrying out a pickup.

\begin{figure}[htb!]
  \vspace{4pt}
\begin{algorithmic}[1]
\Procedure{Main}{}
\Repeat
  \While{robot in visible region}
    \State Wait for next CAM1 frame
  \EndWhile
  \State $f \gets \Call{ProposedGrasps}{}$
  \State $m \gets$ latest trained models
  \State $e \gets \Call{Evaluate}{m, f}$ \Comment{Evaluate all proposals}
  \State $p \gets choosemax(e)$ \Comment{Select best one}
  \If{predicted grasp success probability $< 0.1$ and $\mathrm{UniformVariate}() < 0.95$}
  \State \Return \Comment Avoid making too many failures
  \EndIf
  \State Perform pickup $p$ and obtain feedback
  \If{gripper opening sensor detects failure}
  \State $(c_1,\dots,c_k) \gets (0,\dots,0)$
  \State Add $(p, (c_1,\dots,c_k))$ into the training data
  \ElsIf{pick sequence successful until release}
  \State In another thread in the background,
  \State record frames from CAM2 for 5 s,
  \State $(c_1,\dots,c_k) \gets \Call{Result}{\text{recorded frames}}$
  \State Add $(p, (c_1,\dots,c_k))$ into the training data
  \EndIf
\Until forever
\EndProcedure
\end{algorithmic}
\vfill
\begin{algorithmic}[1]
\Procedure{ProposedGrasps}{}
\State $p \gets $ picture from working area RGBD camera
\State $p' \gets trans(p)$ \Comment{\parbox[t]{.5\linewidth}{Transform to orthogonal projection from above}}
\State $h \gets height(p')$ \Comment{Drop RGB}
\State $d \gets \Call{ClosedGrasps}{h}$
\State $d' \gets \Call{RandomWeightedSample}{d, 2000}$
\State $d'' \gets \Call{ApplyOpenings}{d', h}$
\State $f \gets \Call{AddFeatures}{d'', p'}$ \Comment{Add color features}
\State \Return{$f$}
\EndProcedure
\end{algorithmic}
\vfill
\begin{algorithmic}[1]
\Procedure{Evaluate}{$m$, $f=(f_0,\dots,f_{n-1})$}
\State $p\gets\Call{PredictSuccessProbabilities}{m, f}$
\For{$f_i$ in $f$}
\State $c=(c_1,\dots,c_k)\gets\Call{ExpectedColors}{m, f_i}$
\State $\mathrm{target}_i\gets\argmax_j c_j$ \Comment Index of target color
\State $\mathrm{purity}\gets c_{\mathrm{target}_i}/\sum_j c_j$
\State $\mathrm{recovered}\gets c_{\mathrm{target}_i} p_i$ \Comment Expected recovered pixels
\State $v_i\gets\Call{PurityValue}{\mathrm{purity}}\times \mathrm{recovered}$
\EndFor
\State \Return{$f$ with predicted grasp success probabilities given by $p_i$, values given by $v_i$, and target colors given by $\mathrm{target}_i$ for each $f_i$}
\EndProcedure

\end{algorithmic}
\caption{\label{fig:datagen}
  The overall algorithm that generates the learning data set.
  In order to avoid making silly-looking pickups when the working area is empty,
  a selected best grasp that would have a low success probability is skipped
  19 times out of 20.
  All such grasps are not skipped to ensure the training process
  does not come to a standstill.
 See Fig.~\ref{fig:cam2result} for the \textsc{Result} procedure.
}
\end{figure}

\begin{figure}[tb!]
  \centering
  \includegraphics[width=0.9\columnwidth]{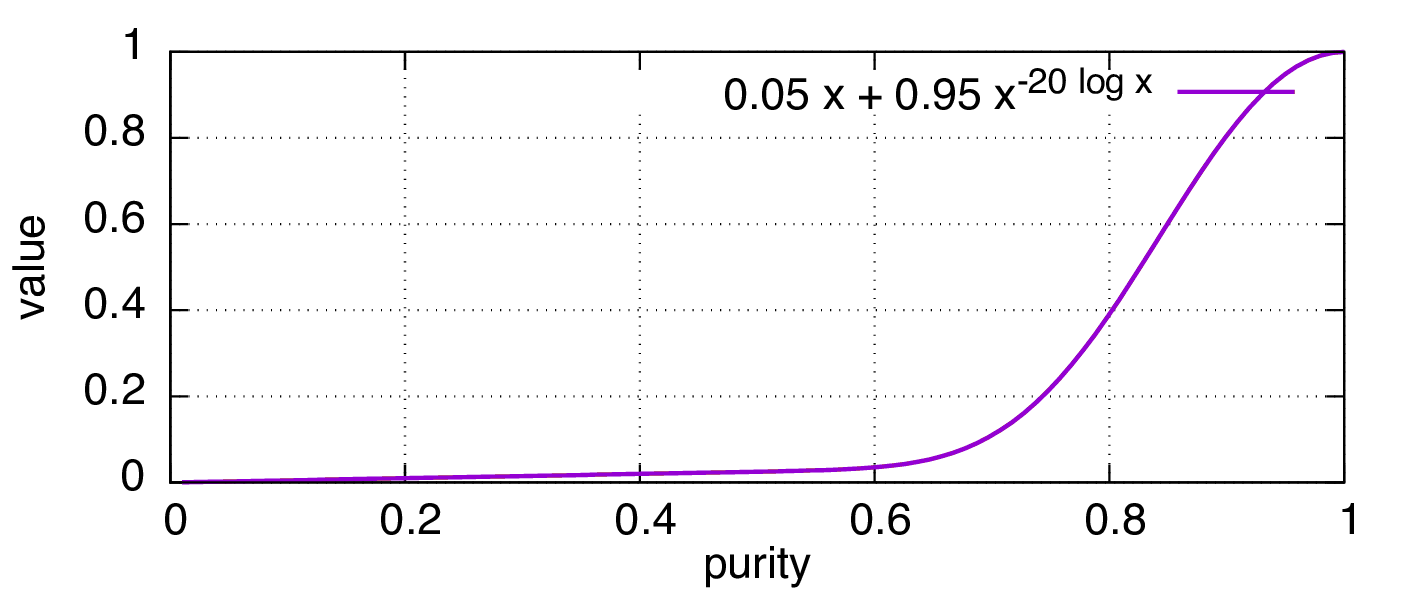}
  \caption{\label{fig:purity_value}The \textsc{PurityValue} function used in grasp evaluation.}
\end{figure}

\begin{figure*}[tb]
  \vspace{5pt}
\begin{algorithmic}[1]
\Procedure{Result}{CAM2 frames}
\State Calculate background level for each pixel as the 20th percentile of the pixel's depth values over time (this is a stable estimate unaffected by objects that are visible at each pixel in only few of the frames)
\State For each frame, form foreground mask as pixels that are at least 6 mm closer than the background level
\State Within a hand-specified region of interest, calculate, for each frame, the
 volume above background within the foreground mask
\State In order to reduce noise, apply minimum filter over time with window of 9 and choose the best frame by maximum filtered volume.
\State \Return counts $(c_1,\dots,c_k)$ of different target colors within the foregound mask (different target colors are defined simply by rectangular boxes in the HSV color space)
\EndProcedure
\end{algorithmic}
\caption{\label{fig:cam2result}The CAM2 result processing algorithm.}
\end{figure*}

\begin{figure}
\begin{centering}
a)
\includegraphics[width=3cm]{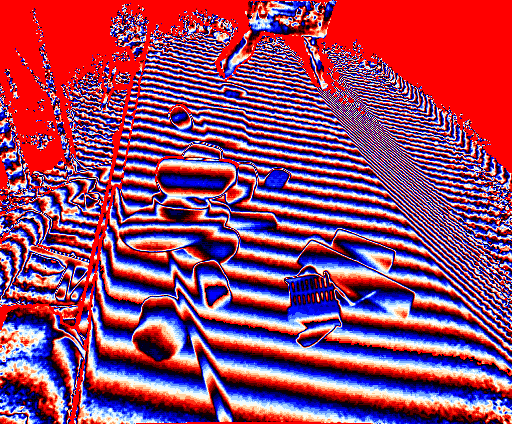}
\includegraphics[width=3cm]{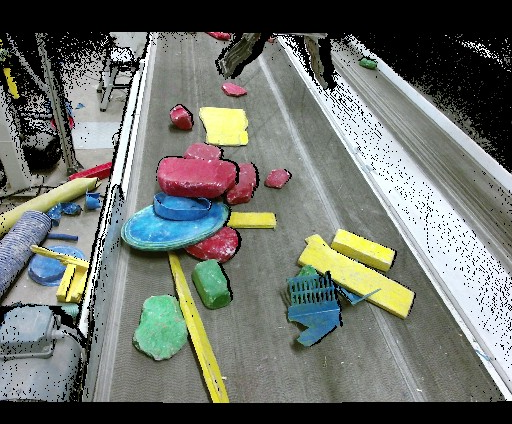}\\
b)
\includegraphics[width=3cm]{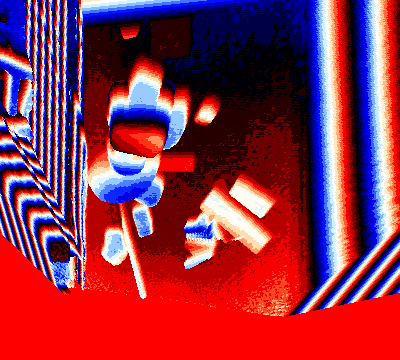}
\includegraphics[width=3cm]{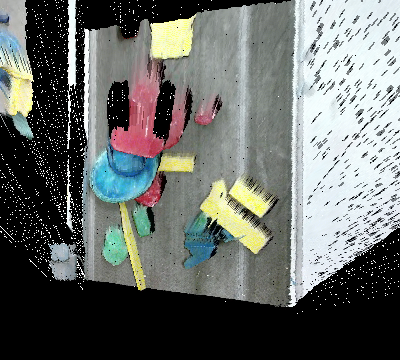}\\
c)
\includegraphics[width=3cm]{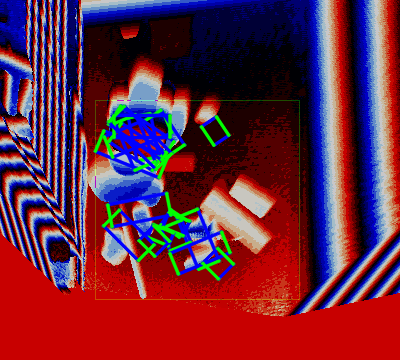}
\includegraphics[width=3cm]{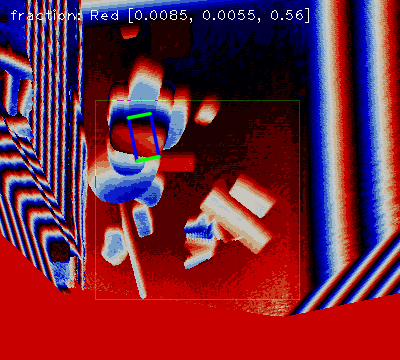}\\
d)
\includegraphics[width=3cm]{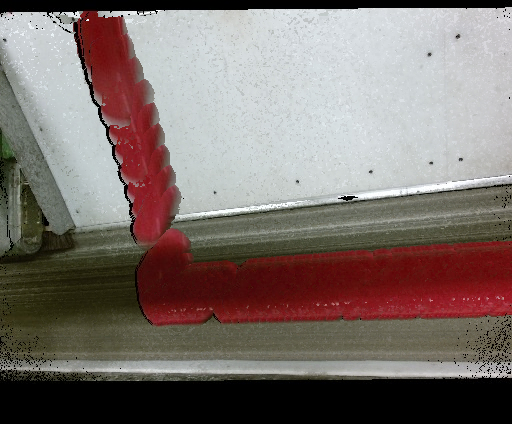}
\includegraphics[width=3cm]{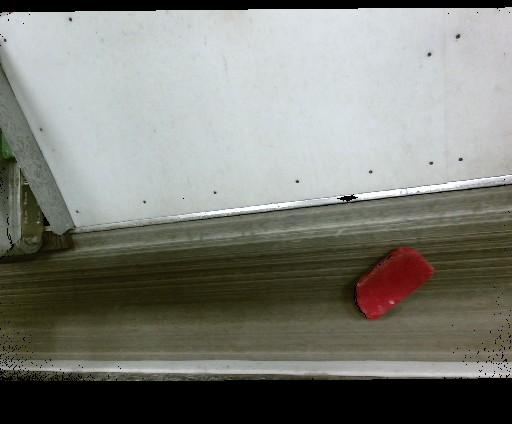}\\
e)
\includegraphics[width=3cm]{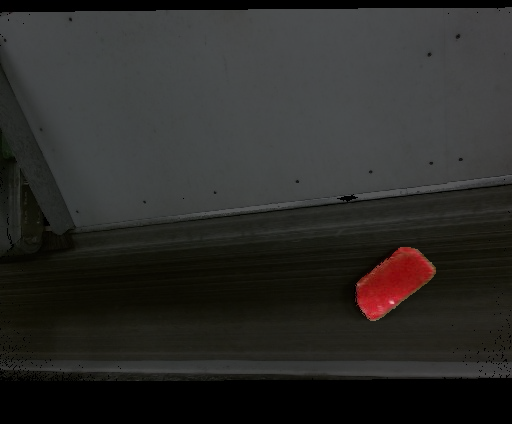}
\includegraphics[width=3cm]{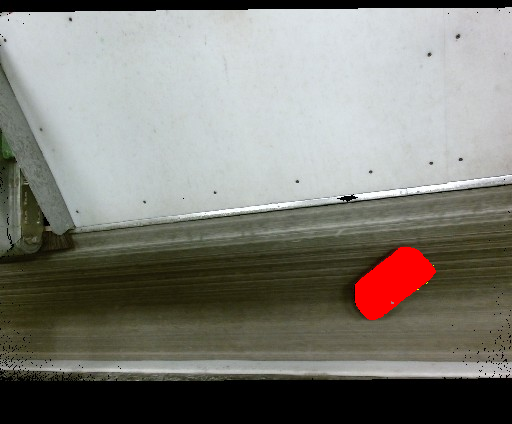}\\
\end{centering}
\caption{\label{fig:pipelineimages}
Images from various points of the software pipeline. Best
viewed in color.
a) The CAM1, depth and RGB.
b) The CAM1, projected to be seen from above as a heightmap.
c) Grasp locations.
Left: a sample of grasps generated by the fixed-function first stage.
This sample is of size 20; the one used by the algorithm is size 2,000.
Right: the grasp
to be executed by the system,
chosen by the machine learning algorithms.
d) CAM2 RGB images after the robot has executed the grasp and throw.
Left: Frames from the output camera, superimposed. Right: the selected
frame (see Fig.~\ref{fig:cam2result}).
e) Processed output camera images.
Left: fg-bg segmentation based on CAM2 depth, with background darkened
Right: The fixed-function color decisions made by the system for the
output image, superimposed on the original image.
This is the feedback based on which the machine learning
system learns to make pure pickups (in this case that all of the picked
object was red).}
\end{figure}
\begin{figure}[tb!]
\begin{centering}
a)
\includegraphics[width=2cm]{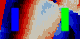}\\
b)
\includegraphics[width=2cm]{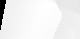}
\includegraphics[width=2cm]{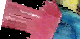}
\includegraphics[width=2cm]{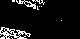}\\
c)
\includegraphics[width=2cm]{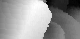}
\includegraphics[width=2cm]{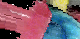}
\includegraphics[width=2cm]{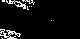}\\
d)
\includegraphics[width=2cm]{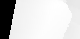}
\includegraphics[width=2cm]{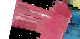}
\includegraphics[width=2cm]{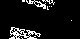}\\
\caption{\label{fig:featureimages}
Data used in grasp selection.
To make the learning problem easy, we give images
centered on the gripper and both fingers of the gripper.
Before being given to the machine learning algorithm, these
images are shrunk as described in the text.
a) A debug image of center images, showing the gripper fingers
superimposed on the heightmap with a colormap.
b) The center images: heightmap, RGB and unknown map (i.e., which
pixels are invisible in the projected heightmap.
c) The left finger images, similar.
d) The right finger images, similar.
}
\end{centering}
\end{figure}

\subsubsection{Heightmap projection}

We project the RGBD camera image into a heightmap with 5 mm pixel resolution,
with orthogonal projection
by rendering it through the OpenGL 3D API into a buffer (see Fig.~\ref{fig:pipelineimages}).
In order to avoid colliding to occluded objects,
the projection code marks pixels
that are occluded by objects to their maximum possible heights and
additionally generates a mask indicating such unknown pixels.
This is accomplished by rendering a frustum for each pixel, the sides
of the frustum being rendered with a special ``unknown'' color
if the D coordinate difference between the pixel and its neighbour exceeds
a certain limit.

\subsubsection{Fixed-function first stage grasp finding}

\begin{figure*}
  \vspace{5pt}
  \begin{algorithmic}[1]
    \Procedure{ClosedGrasps}{$h$, $\mathrm{num\_angles}$, $\mathrm{finger\_thickness}$, $\mathrm{finger\_width}$, $\mathrm{min\_opening}$, $\mathrm{max\_opening}$}
    \State $\mathrm{res}\gets\mathrm{new\ List}$
    \For{$\alpha\gets0,(1/\mathrm{num\_angles})\pi,\dots,((\mathrm{num\_angles}-1)/\mathrm{num\_angles})\pi$}
    \State $h'\gets \mathrm{maximum\_filter}( \mathrm{rotate}(h, \alpha), (\lceil\mathrm{finger\_thickness}\rceil, \lceil\mathrm{finger\_width}\rceil))$
    \ForAll{rows $y$}
    \State $g\gets\Call{ClosedGrasps1D}{h'[\cdot, y],
      \mathrm{min\_opening} + \mathrm{finger\_thickness},
      \mathrm{max\_opening} + \mathrm{finger\_thickness}}$
    \State Interpret each $(x_0,x_1,z,v)$ in $g$ as the 3D grasp rectangle
    \State $[x_0+\mathrm{finger\_thickness}/2,x_1-\mathrm{finger\_thickness}/2]\times[y-\mathrm{finger\_width}/2,y+\mathrm{finger\_width}/2]\times\{z\}$,
    \State rotate it by $-\alpha$ about the center of $h'$, and append to $\mathrm{res}$ together with the value $v$ 
    \EndFor
    \EndFor
    \State \Return{$\mathrm{res}$}
    \EndProcedure
\end{algorithmic}
  \caption{\label{fig:graspfindingalg}
Exhaustive closed grasp finding algorithm; all sizes are in pixels; the algorithm runs in
    time $\mathcal{O}(\mathrm{num\_angles}\times\mathrm{heightmap\ size})$.  See Fig.~\ref{fig:graspfinding} for the overall logic. This algorithm uses the one-dimensional
grasp finding algorithm defined in Fig.~\ref{fig:graspfindingalg1d}.
}
\end{figure*}
\begin{figure}[tb!]
\begin{algorithmic}[1]
    \Procedure{ClosedGrasps1D}{\mbox{$h[0,\dots,n\!-\!1],d_{\min},d_{\max}$}}
    \State $\mathrm{res}\gets\mathrm{new\ List}$
    \State $\mathrm{stack}\gets\mathrm{new\ Stack}$
    \For{$i\gets 1,\dots,n-1$}
      \If{$h[i] > h[i-1]$}
        \State $\mathrm{stack.push}(i - 1)$
      \ElsIf{$h[i] < h[i-1]$}
        \While{$\mathrm{stack.size}() > 0$}
          \State $\mathrm{top}\gets \mathrm{stack.top}()$
          \If{$d_{\min} \le i - \mathrm{top} \le d_{\max}$}
          \State $z\gets \max(h[\mathrm{top}], h[i])$
          \State $v\gets h[\mathrm{top}\!+\!1] - h[\mathrm{top}] + h[i\!-\!1] - h[i]$
          \State $\mathrm{res.append}((\mathrm{top}, i, z, v))$
          \EndIf
          \If{$h[i] > h[\mathrm{top}]$}
            \State \textbf{break}
          \EndIf
          \State $\mathrm{stack.pop}()$
        \EndWhile
      \EndIf
    \EndFor
    \State \Return{$\mathrm{res}$}
    \EndProcedure
    
  \end{algorithmic}
  
  \caption{\label{fig:graspfindingalg1d}
    The stack-based 1D closed grasp finding algorithm. Given a 1D array $h$, the procedure returns all quadruples $(i_0, i_1, z, v)$ such that $d_{\min} \le i_1 - i_0 \le d_{\max}$ and $h[i]>z=\max\{h[i_0],h[i_1]\}$ for all $i_0 < i < i_1$, where $v$ is a rudimentary metric of the quality of the grasp used for weighting the initial random sample in \textsc{ProposedGrasps}.  The stack holds rising steps of the height curve and when the height steps down, grasps are formed by popping left sides from stack until the left side is below the right side.
  }
\end{figure}

\begin{figure}[tb!]
  \centering
  {\includegraphics[width=\columnwidth,trim=0px 15px 0px 0px]{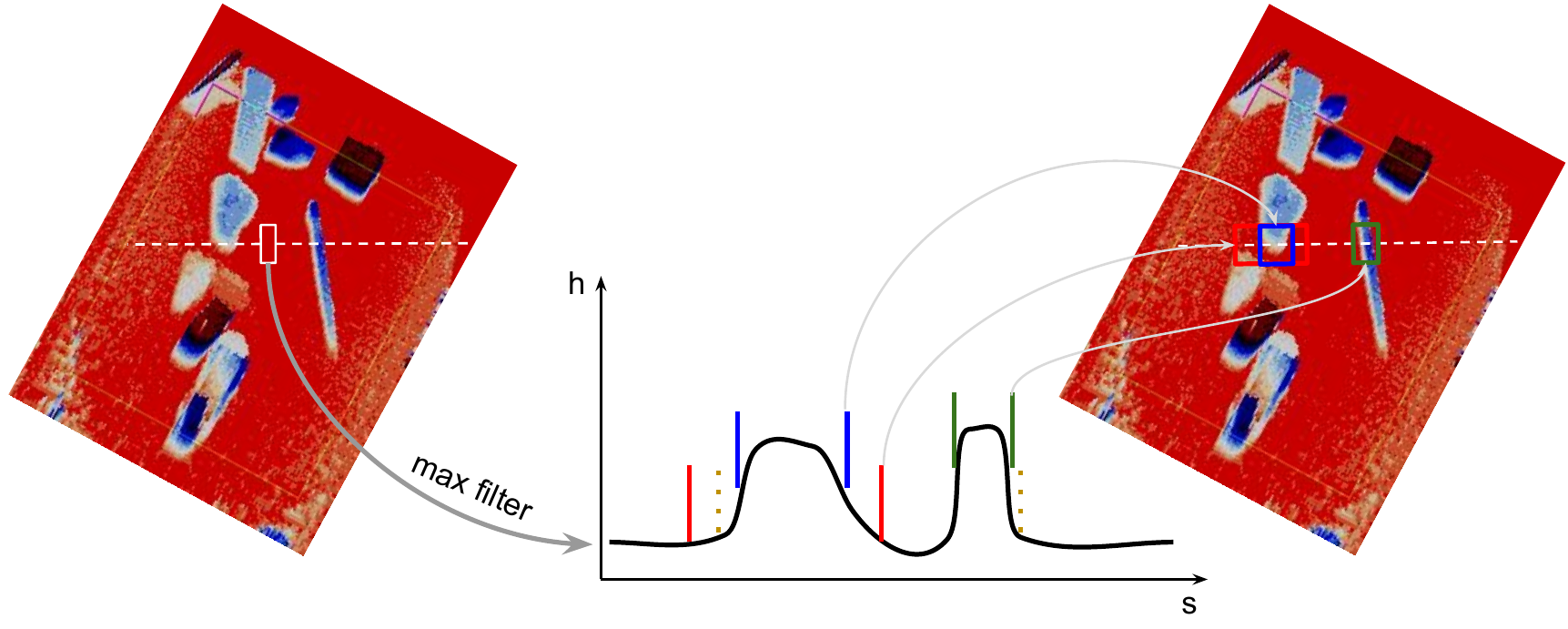}}
  \caption{\label{fig:graspfinding}Exhaustive search of closed grasps: a rectangular kernel
    of the shape of the gripper finger is moved across a line on the
    heightmap and yields (by maximum-filtering) a height curve $h(s)$
    indicating the minimum possible height of the gripper finger given
    the conveyor contents; closed grasps aligned on the line are
    determined by pairs $(s_0, s_1)$ such that $h(s_0) < h(s) >
    h(s_1)$ for all $s_0 < s < s_1$ (three examples are shown in the
    figure). The dotted brown lines indicate a grasp that is not included in the set of 
    closed grasps since the curve
    drops below its values at the sides at some point in the middle
    (i.e., condition $h(s_0) < h(s) > h(s_1)$ is not satisfied).
    Excluding such grasps makes all overlapping grasps on the line nested,
    which allows a stack-based algorithm (shown in Fig.~\ref{fig:graspfindingalg1d})
    to generate all closed grasps
    in linear time w.r.t. the number of pixels on
    the line.  We use 16 discrete directions for the search line, which is implemented
    in practice
    by scanning all rows of a rotated heightmap as shown in the figure.
  }
\end{figure}

Possible grasps are modeled using a rectangle representation similar to those used in
\cite{pinto16supersizing, jiang2011efficient}.
The left side of a grasp rectangle specifies the position and the extent of
the inner side of the left gripper finger, and the right side specifies the position
and the extent of the inner side of the right gripper finger.  The width of the rectangle
specifies the gripper opening. The rectangle also has a $z$ coordinate, specifying
the height at which to grasp (not visible in figures).

The possible grasps are generated starting from an exhaustive search of so
called \emph{closed grasps}, grasps in which the fingers of the gripper touch the heightmap
from both sides and the heightmap rises between the two points, see Fig.~\ref{fig:graspfinding}.
A random sample of closed grasps, weighted by a rudimentary metric of grasp quality
(see Fig.~\ref{fig:graspfindingalg1d}), is generated for further evaluation.

The \textsc{ApplyOpenings} procedure duplicates each closed grasp for all possible extra openings allowed by the heightmap. It uses a geometric model of the nonlinear opening movement of the gripper and reads the heightmap to determine how much the gripper can be opened from the original closed grasp position without colliding with the heightmap. The $z$-coordinate is increased if necessary, but only if the grasp remains a closed grasp (i.e., the inner sides of the gripper fingers touch the heightmap from both sides at the closed grasp position).

\subsubsection{Choice of grasp}

As discussed in Section~\ref{sec:formalwastesorting}, the system
is solving a multi-objective optimization problem.
As the single objective to optimize,
we define a slightly ad hoc utility function as the product of the expected amount
of correct color in the target area, multiplied by a nonlinear
function of purity (see Fig.~\ref{fig:purity_value}).
This nonlinear function strongly prefers grasps with expected 
purity above 80\% (at a real site, this threshold would probably be made significantly higher)
and multiplying by the expected recovery
avoids making very narrow picks that are likely to fail but would yield high purity
if they succeeded.

\subsubsection{Model learning}

Parallel to the main sorting loop, the model learning loop is run constantly.
In the very beginning, without any data, a ``null model'' is produced, which
predicts 1.0 grasp success probability for all grasps and estimates the output to be
a constant number of pixels of a special ``unknown'' color.


On each iteration, two models are trained from scratch: one classifier model
to predict the probability of success of
a pickup, and one regression model to predict class proportions that end
up in the drop zone (for successful pickups only).
All training data is used on every iteration.
For both models, we used Extremely Randomized Trees \cite{Geurts:2006:ERT:1132034.1132040}
as implemented by Scikit-learn \cite{scikit-learn}.
This algorithm was chosen because they rarely overfit
and are fast to train and apply.

For each proposed grasp rectangle, a number of features are gathered for machine learning.
A portion of the RGBD heightmap is rotated to the grasp rectangle.
Different features are used for the two models. 
For grasp success probability model (see Fig.~\ref{fig:featureimages}):
\begin{itemize}
  \item
  $80\times39$ pixel ($40\times19.5$~cm) slices of the heightmap, RGB image, and unknown mask aligned at the left
    finger, center, and right finger of the gripper (including a margin of 4~cm around the grasp rectangle), downscaled by a factor of four in both directions,
  \item the opening of the grasp and extra opening to be applied when grasping, and
  \item the height of the grasp (which is also subtracted from the
    heightmap slices so as to yield translation invariant features), and
  \item the position and angle of the grasp rectangle on heightmap (to allow learning, e.g., boundary effects on the work area).
\end{itemize}
For the color model, the image features are replaced by fewer and more downscaled ones:
\begin{itemize}
  \item $80\times39$ pixel ($40\times19.5$~cm) slices of the heightmap and RGB image aligned at the 
    center of the gripper, downscaled by a factor of 8 in both directions.
  \item the opening, height and position features as above.
\end{itemize}

The training data includes the above features for each attempted pick, and
the result is given by the numbers of pixels $(c_1,\dots,c_k)$ of each target color.

The success probability model is trained to predict the grasp result, which is 0, if $c_1=\dots=c_k=0$ and 1, otherwise.

The color model is only trained on successful grasps.
It is trained to estimate the expected amounts $(c_1,\dots,c_k)$ of different colors in the target area.
However, the system worked significantly better in practice, when, instead of absolute pixel counts,
we trained the system on proportions of different colors within the foreground mask of the target area.
This normalization was used in the experiment.

\subsection{Procedure}

The experiment was run in eight half hour parts. 
CAM1 was calibrated in the beginning and in the middle of the experiment.

1,743 pickups were carried out, and the model was allowed to learn the whole time.
Conveyor 1 (see Fig.~\ref{fig:hw}) was controlled manually in small steps 
so as to let the robot clear the piles. Otherwise the system worked autonomously while it was running.
Conveyor 2 was run constantly and Conveyor 3 was stopped so as to keep the drop
zone clear for recording feedback.  Conveyor 3 was cleared from any missed objects during experiment breaks,
and occasionally the system was stopped for removing any stuck objects between the belts.


\section{RESULTS}

\begin{figure}[tb!]
  \centering
  \includegraphics[width=0.494\columnwidth]{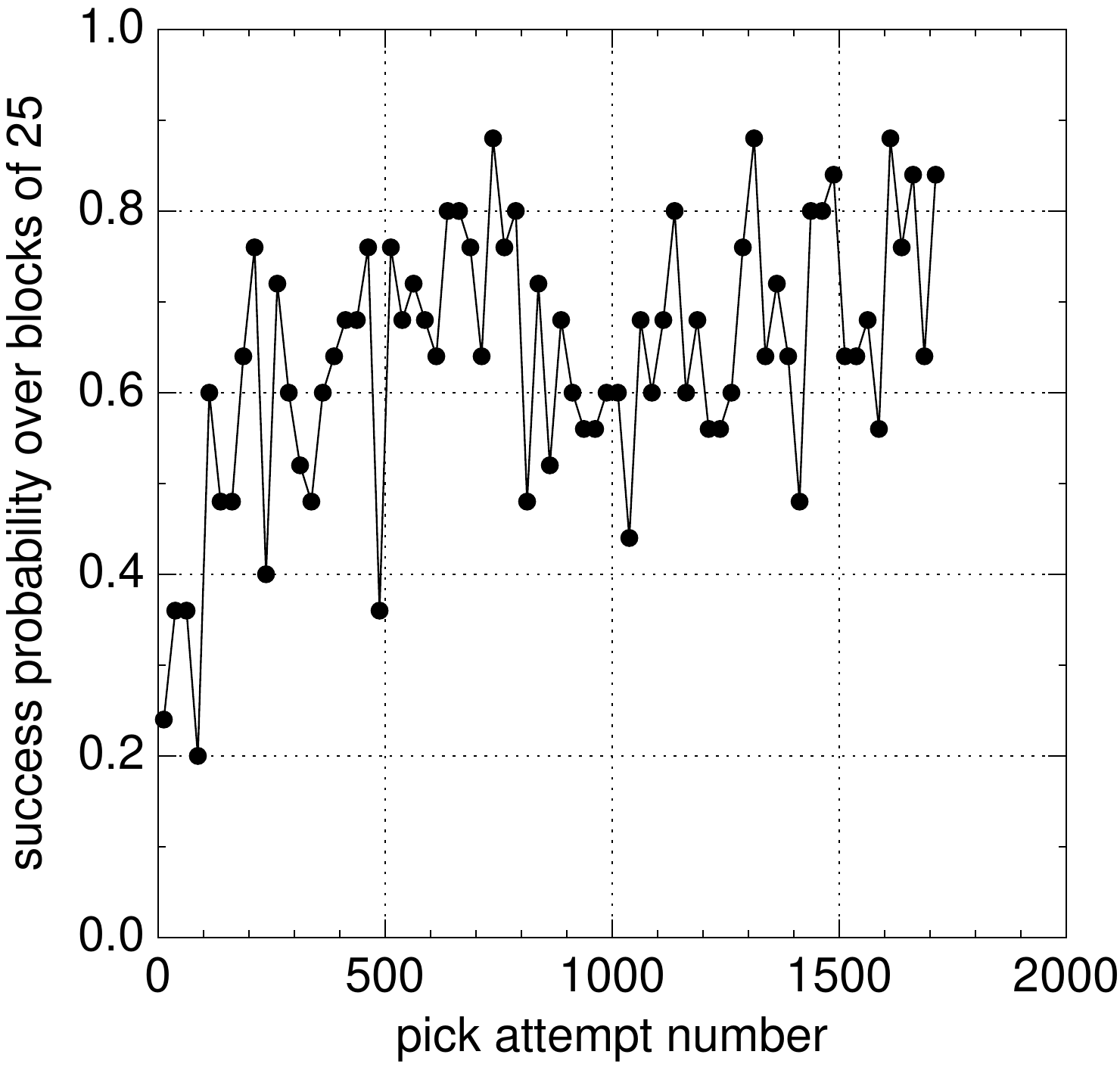}%
  \includegraphics[width=0.494\columnwidth]{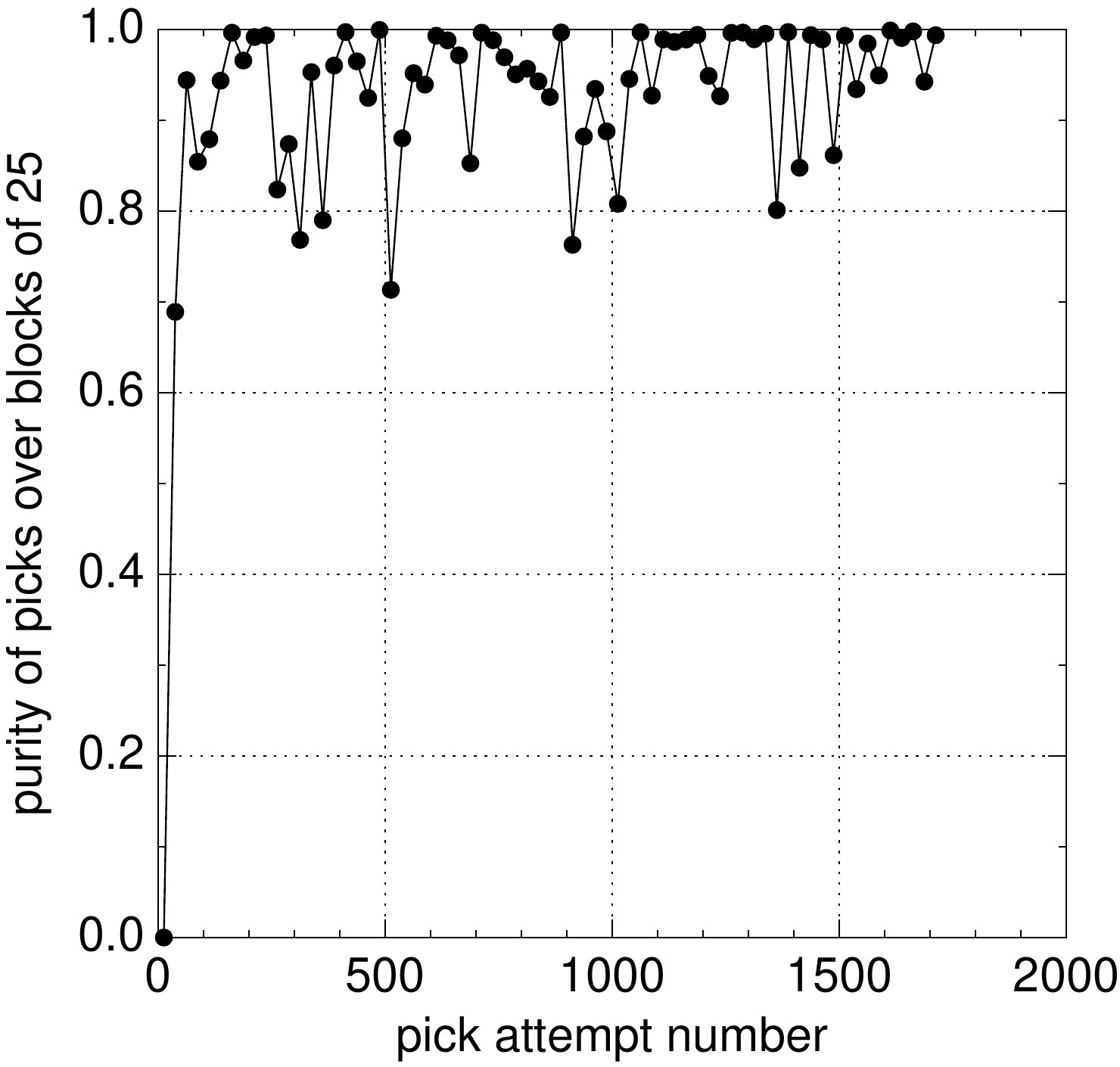}
  \caption{\label{fig:results}Graphs illustrating the success probability of the grasps and the overall purity of the picks over time in the experiment.}
\end{figure}

\begin{figure}[tb!]
\begin{centering}
  \vspace{5pt}
  \includegraphics[width=6.2cm]{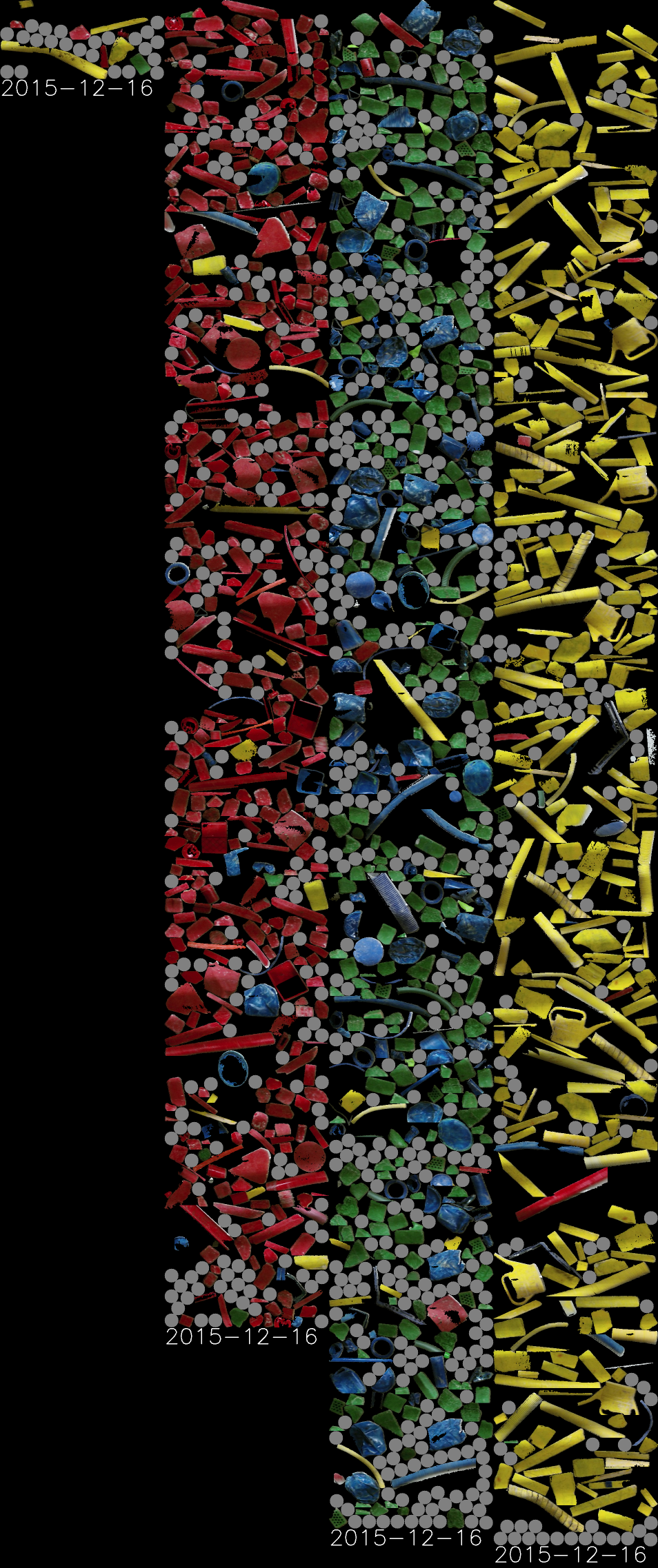}
  \caption{\label{fig:chutes}Visualization of the sorting result over the whole experiment: CAM2 foreground segment images from objects thrown are stacked for picks with unknown, red, blue-green, and yellow target fraction; time goes up in each column.  The time in each column is independent. Gray circles indicate failed picks.}
\end{centering}
\end{figure}

The system learned quickly to grasp the objects and sort them by color.
Figure~\ref{fig:results} shows the success probability of the grasps over blocks of 25 trials,
as well as the overall purity of the picks (total number of pixels of the target fraction divided
by the total number pixels seen in the drop zone) over blocks of 25 trials.  Figure~\ref{fig:chutes}
shows a visualization of the sorting result over the whole experiment.
Representative example grasps are shown in Fig.~\ref{fig:representative-picks}.

The cycle time (between two consecutive pickups)
was about 8 s in total, including approximately 2--3 s of computation
for generating the best grasp from the CAM1 image.


%
%
%
%
%

\renewcommand{\tabcolsep}{1pt}
\newlength{\picw}
\setlength{\picw}{3.38cm}
\begin{figure*}[tb!]
\centering
  \vspace{5pt}
\begin{tabular}{lcccccc}
a)&
        \includegraphics[width=\picw]{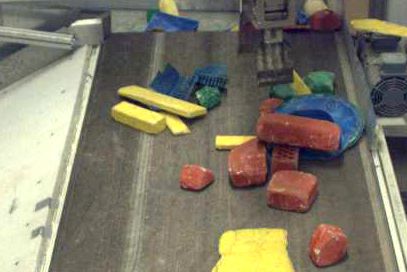}&
        \includegraphics[width=\picw]{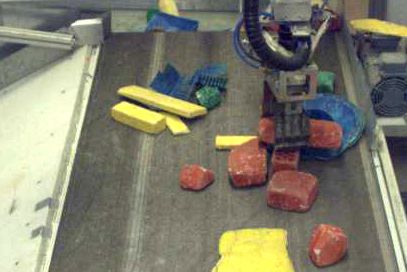}&
        \includegraphics[width=\picw]{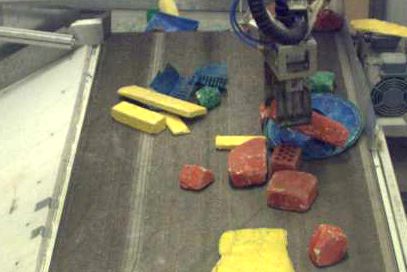}&
        \includegraphics[width=\picw]{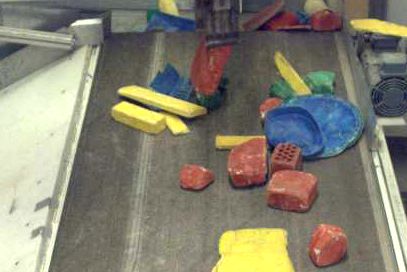}&
        \includegraphics[width=\picw]{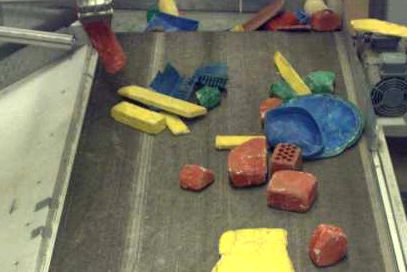}&
        \\
&
        \includegraphics[width=\picw]{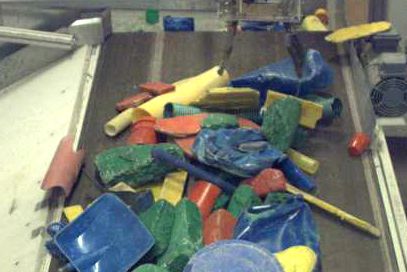}&
        \includegraphics[width=\picw]{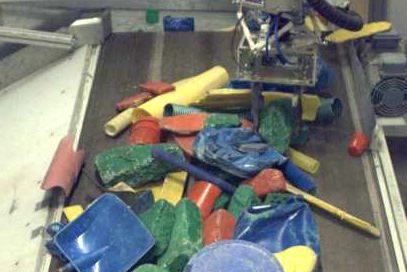}&
        \includegraphics[width=\picw]{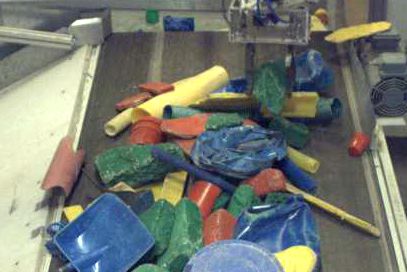}&
        \includegraphics[width=\picw]{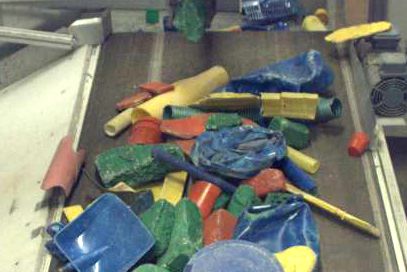}&
        \includegraphics[width=\picw]{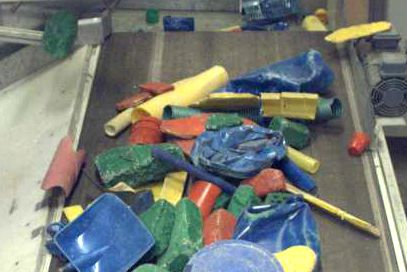}&
        \\
&
        \includegraphics[width=\picw]{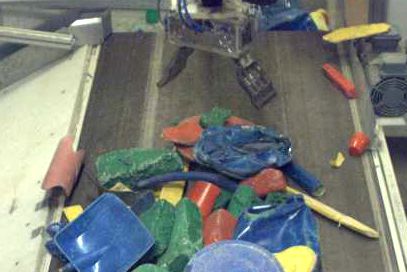}&
        \includegraphics[width=\picw]{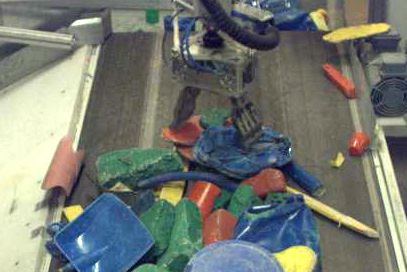}&
        \includegraphics[width=\picw]{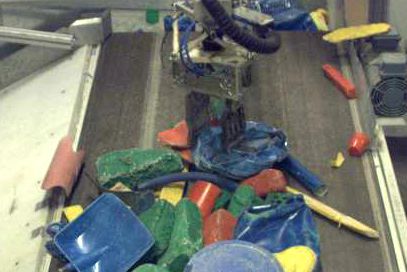}&
        \includegraphics[width=\picw]{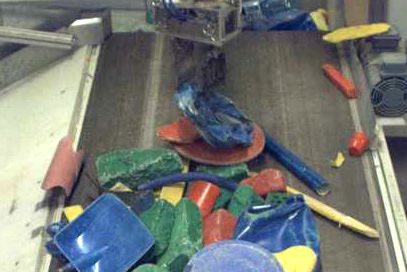}&
        \includegraphics[width=\picw]{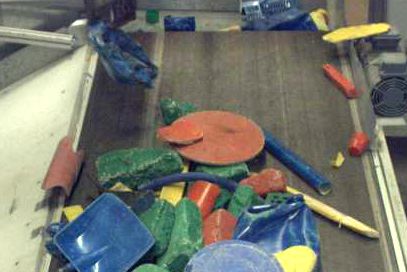}&
        \\
&
        \includegraphics[width=\picw]{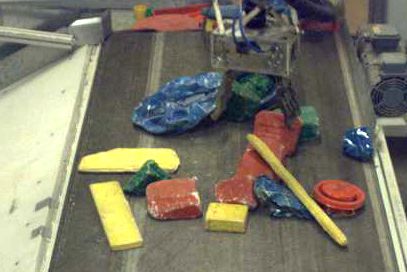}&
        \includegraphics[width=\picw]{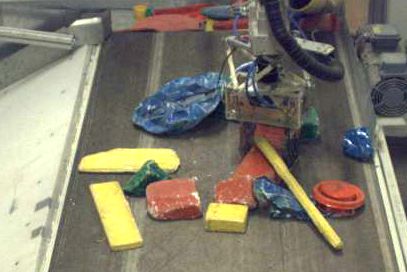}&
        \includegraphics[width=\picw]{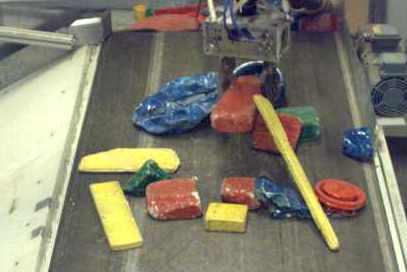}&
        \includegraphics[width=\picw]{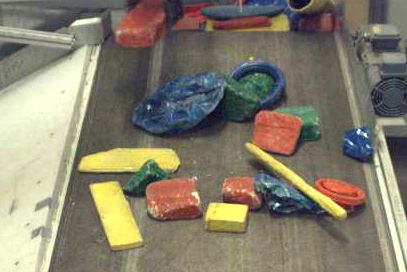}&
        \includegraphics[width=\picw]{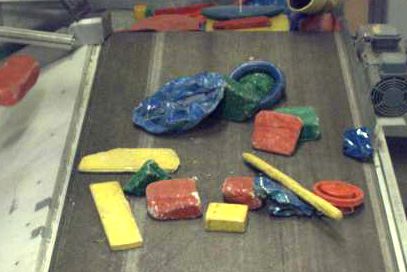}&
        \\
b)&
        \includegraphics[width=\picw]{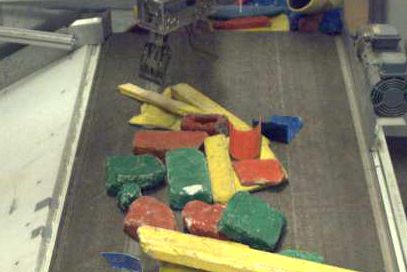}&
        \includegraphics[width=\picw]{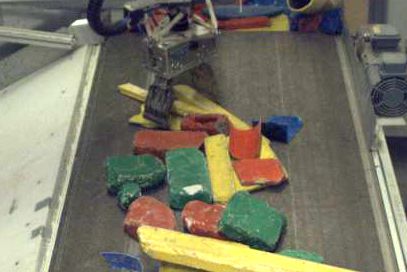}&
        \includegraphics[width=\picw]{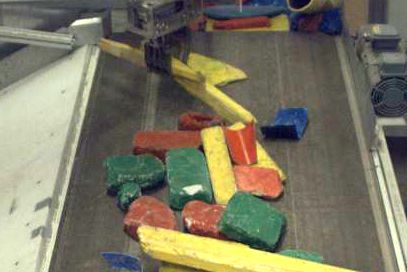}&
        \includegraphics[width=\picw]{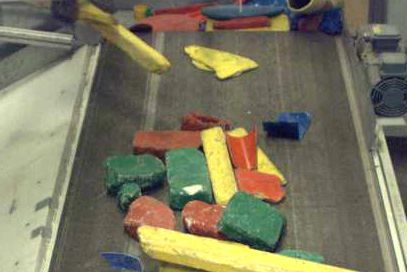}&
        \includegraphics[width=\picw]{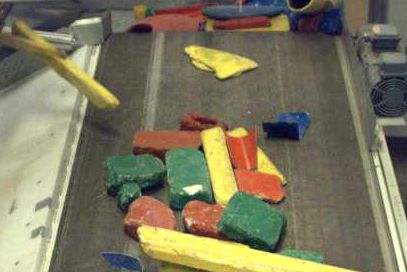}&
        \\
c)&
        \includegraphics[width=\picw]{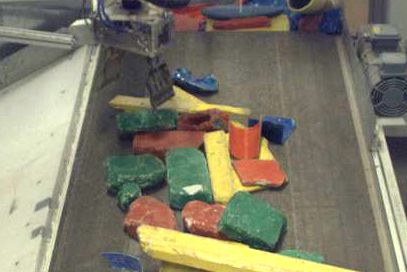}&
        \includegraphics[width=\picw]{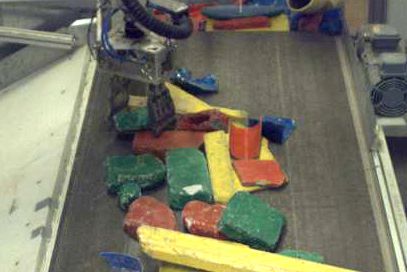}&
        \includegraphics[width=\picw]{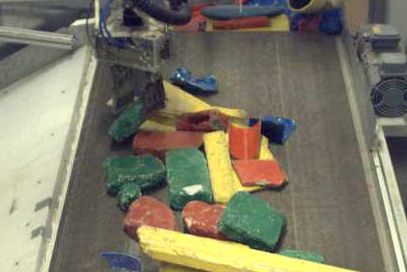}&
        \includegraphics[width=\picw]{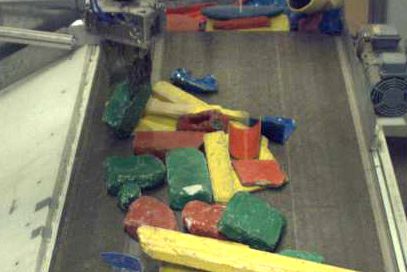}&
        \includegraphics[width=\picw]{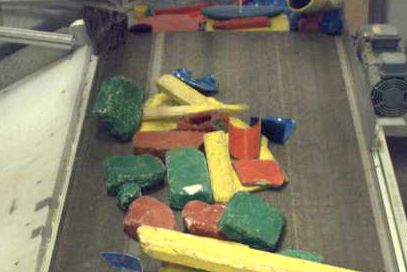}&
        \\
d)&
        \includegraphics[width=\picw]{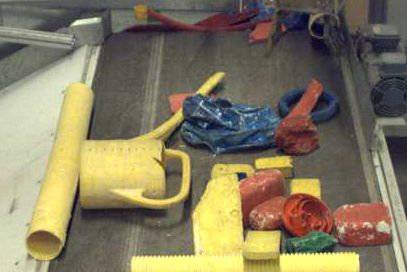}&
        \includegraphics[width=\picw]{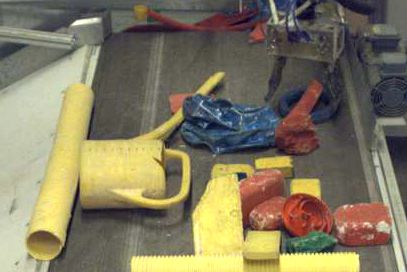}&
        \includegraphics[width=\picw]{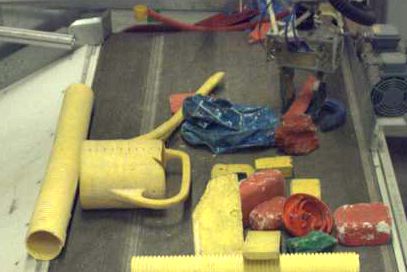}&
        \includegraphics[width=\picw]{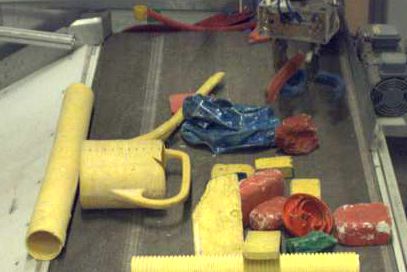}&
        \includegraphics[width=\picw]{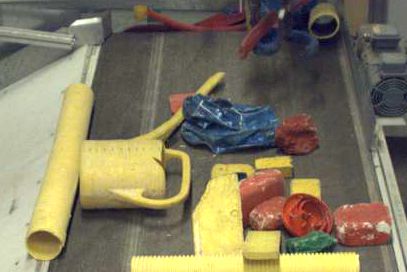}&
        \\
e)&
        \includegraphics[width=\picw]{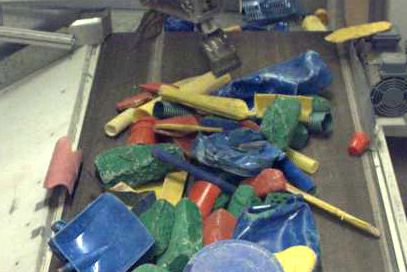}&
        \includegraphics[width=\picw]{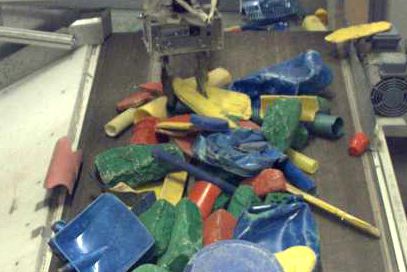}&
        \includegraphics[width=\picw]{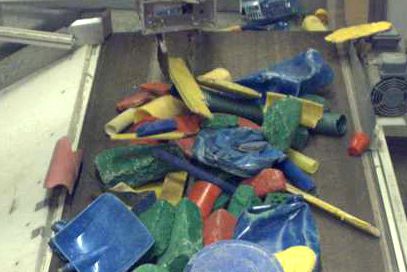}&
        \includegraphics[width=\picw]{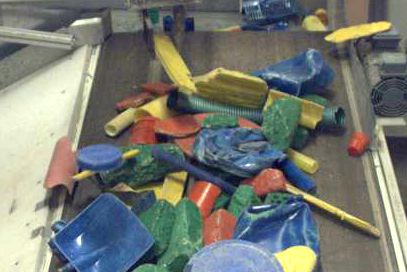}&
        \includegraphics[width=\picw]{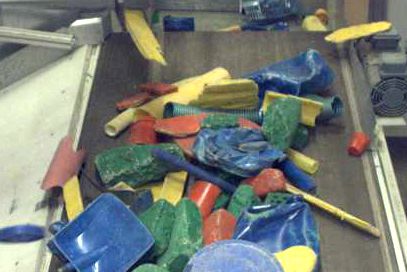}&
\end{tabular}
\caption{\label{fig:representative-picks}
Some representative example grasps by the system. Best viewed in color.
Frames of video chosen for clarity.
a) Successful picks that deposited one object of the predicted color
to the output area.
b) A successful pick that picked more than one object of the correct color.
c) A failed pickup where the object slipped from the grasp of the system
and the gripper opening sensor noticed this.
d) A failed pickup that accidentally
picked, in addition to the red object (the predicted
color), the blue object underneath.
e) A special failed pick where the feedback system did not see the blue
object that flew to the second conveyor outside the CAM2 visible area.
Not shown: some pickups, especially early in the learning process, predicted
the wrong color for the picked up object.
}
\end{figure*}

\section{DISCUSSION AND FUTURE WORK}

We have demonstrated a system that learns to sort
a densely cluttered pile of objects
by class.

The system is general and should be able
to learn just about any sorting task, provided it is run long
enough.
The only theoretical limitation we see for the proposed system
is that it might be able to learn to fool the feedback processing.
For example,
the system could learn to throw multiple objects so that only the object
of the right class is seen (e.g., it ends up on top of the others).
In this case, the system would be making many
impure throws.
This behaviour was not observed in our experiment
but might appear in a more complex task. The system can only be as good as the
feedback it gets. Possible ways of alleviating this problem if it were to occur
are, e.g.,
taking feedback pictures while the objects are flying through the air or having
the robot manipulate the thrown objects more to ensure there are only ones from the correct class.


The practical demonstration of the proposed system is still relatively
limited.
For example,
the region around a grasp that the system
sees as features is small due to performance reasons;
this prevents it from learning to not pick the end
of a plank that is under a pile of other objects
(except indirectly
by learning that such pickups sometimes
cause more problems than ones at larger height
from the belt).
This limitation is not inherent in our architecture:
replacing the learning component with a more powerful
one (e.g., one using deep learning) and increasing the input region
size will likely make handling this situation possible.

Another possible practical limitation stems from the fact that
the system has to re-learn the classification of objects on the working area.
While being also a strength (objects might look different in the
clutter of the
working area), this may make learning slow if the classes are difficult.
This can be alleviated in several ways, such as using the results
of existing classifiers as input features on the working area, which
might work, especially if the regression model is biased to be symmetric
with respect to change of classes.

Next steps of future work include sorting objects based on
classes not determined by color as well as systems that
specifically aim at picking more than one object at a time.

\section{ACKNOWLEDGMENTS}

The authors would like to thank the ZenRobotics research assistants, especially Risto Sirvi{\"o} for supervising many of the experiments and Risto Sirvi{\"o} and Sara Vogt for annotating experiment data.  The authors would also like to thank Risto Bruun, Antti Lappalainen, Arto Liuha, and Ronald Tammep{\~o}ld for discussions and PLC work, Matti K\"a\"ari\"ainen, Timo Tossavainen, and Olli-Pekka Kahilakoski for many discussions, and Risto Bruun, Juha Koivisto, and Jari Siitari for hardware work.  This work also makes use of the contributions of the whole ZenRobotics team through the parts of our product that were reused in this prototype.

\end{document}